\documentclass{article}

\usepackage{PRIMEarxiv}
\usepackage[utf8]{inputenc}
\usepackage[T1]{fontenc}
\usepackage{hyperref}
\usepackage{url}
\usepackage{booktabs}
\usepackage{amsmath}
\usepackage{amsfonts}
\usepackage{amssymb}
\usepackage{array}
\usepackage{multirow}
\usepackage{microtype}
\usepackage{fancyhdr}
\usepackage{graphicx}
\graphicspath{{media/}}

\title{An Ensembled Latent Factor Model via Differential Evolution and Gradient Descent Optimization}

\author{
  Rui Zhang \\
  Chongqing Academy of Economics Research \\
  Chongqing, China \\
  \texttt{50261231@qq.com} \\
  \And
  Jinhang Liu \\
  College of Computer and Information Science, Southwest University \\
  Chongqing, China \\
  \texttt{tesla369@email.swu.edu.cn} \\
  \And
  Wenbo Zhang$^{*}$ \\
  College of Computer and Information Science, Southwest University \\
  Chongqing, China \\
  \texttt{z13996091260@email.swu.edu.cn}
}

\begin{document}
\maketitle

\begin{abstract}
High-dimensional and incomplete (HDI) data are prevalent in many real-world big data scenarios. Latent factor models serve as a common representation learning approach, capable of uncovering informative latent factors from such data. Nevertheless, most existing latent factor models rely solely on gradient descent for optimization, which may lead to insufficient and biased representations, particularly when dealing with heterogeneous HDI data. Thus, this study proposes an Ensembled Latent Factor Model via Differential Evolution and Gradient Descent Optimization (ELFM-DEGDO) with two-fold designed: 1) two diverse latent factor models are independently modeled via differential evolution and gradient descent optimization, respectively, and 2) the two diverse latent factor models are combined via a customized self-adaptive weighting mechanism to effectively fuse their strengths. By leveraging the complementary advantages of both optimization paradigms, ELFM-DEGDO is able to produce more comprehensive and less biased representations for HDI data. Three HDI datasets are tested to show that ELFM-DEGDO consistently performs better than related several latent factor models.
\end{abstract}


\section{Introduction}

Matrices are a fundamental tool for modeling pairwise interactions among entities \cite{ref1,ref2,ref3}. A representative example is the user-item matrix, which is extensively used to characterize online e-commerce systems \cite{ref20,ref21,ref22,ref23}. In this matrix, rows and columns denote items and users, respectively, while each element captures the degree of a user's interest or preference in a particular item \cite{ref24,ref25,ref26}. More generally, matrix representations are ubiquitous in both scientific research and industrial practice, supporting a wide range of applications, including Web services \cite{ref4,ref5,ref27,ref28,ref29,ref30}, recommender systems \cite{ref31,ref32,ref33}, mobile sensing networks, and social service networks \cite{ref34,ref35,ref36}. Such representations play a crucial role in describing complex relational data in real-world scenarios. However, matrices obtained from practical applications are often high-dimensional and incomplete (HDI). This is primarily because it is challenging to fully observe all interactions among a large number of entities, resulting in sparse and partially observed data \cite{ref6,ref7}.

Thus, a central challenge in handling HDI data is how to perform effective and efficient representation learning \cite{ref8,ref9}. This issue is both theoretically important and practically evident. For instance, in modern online serving platform, e.g., Walmart the numbers of users and items are extremely large, whereas only a small fraction of user-item interactions are actually observed \cite{ref10,ref11,ref41,ref42,ref43}. Consequently, most entries in such large-scale matrices remain unobserved, even though they potentially contain rich and valuable information, such as users' implicit preferences for items \cite{ref12,ref44,ref45,ref46}. This highlights the importance of learning accurate latent representations from HDI data, which can facilitate the discovery of hidden knowledge embedded in incomplete observations \cite{ref37,ref38,ref39,ref40,ref47,ref48,ref49,ref50,ref51,ref52}.

Latent factor models are among the most widely used approaches for representation learning on HDI data, owing to their strong predictive performance, computational efficiency, and scalability \cite{ref12,ref13}. Compared with deep neural network approaches \cite{ref17,ref18}, including methods such as AutoRec \cite{ref15} and DCCR \cite{ref16}, they generally require less computational cost. They represent the original HDI matrix through the factorization of two low-rank latent feature matrices \cite{ref57,ref58,ref59,ref60}. The latent representations are iteratively optimized by reducing the reconstruction error between observed entries and their predicted values \cite{ref8,ref53,ref54,ref55,ref56}. In this setting, underlying representations of the row and column entities in the original HDI matrix are encoded by two corresponding latent matrices, respectively \cite{ref12,ref61,ref62,ref63,ref64}.

Although latent factor models have achieved notable success, most existing approaches depend exclusively on gradient descent for optimization \cite{ref12,ref65,ref66,ref67}, which can result in incomplete and biased representations, especially when handling heterogeneous HDI data \cite{ref68,ref69,ref70}. To address this issue, an Ensembled Latent Factor Model via Differential Evolution and Gradient Descent Optimization (ELFM-DEGDO) is proposed. ELFM-DEGDO has two-fold ideas. First, two diverse latent factor models are independently modeled via differential evolution and gradient descent optimization, respectively. Second, the two diverse latent factor models are combined via a customized self-adaptive weighting mechanism to effectively fuse their strengths. As such, the complementary advantages of both optimization paradigms are leveraged, enabling ELFM-DEGDO to produce more comprehensive and less biased representations for HDI data. We conclude the contributions of this paper as follows.

\begin{itemize}
\item An ELFM-DEGDO model is proposed. Through the combination of differential evolution and gradient descent strategies, it is capable of extracting accurate latent features from HDI data.
\item Theoretical analyses, as well as algorithm designs, are discussed for the proposed ELFM-DEGDO model.
\item Extensive empirical studies are tested to evaluate the proposed ELFM-DEGDO.
\item Results obtained from experiments on three real HDI datasets show that ELFM-DEGDO performs more effectively than multiple advanced latent factor models.
\end{itemize}

\section{The Proposed Model}
\subsection{A Latent Factor Model}
Generally, let $R\in\mathbb{R}^{|U|\times |I|}$ denote an HDI matrix. A latent factor model is to learn $X\in\mathbb{R}^{|U|\times f}$ and $Y\in\mathbb{R}^{|I|\times f}$ for making $R$'s rank-$f$ approximation by $\hat{R}=XY^{T}$ based on $R_K$, where $f\ll\min\{|U|, |I|\}$ and $R_K$ denotes $R$'s known entry set \cite{ref7,ref8}.

To obtain $X$ and $Y$, it is essential to define an objective function over $R_K$. This function is typically formulated as a combination of a loss term and a regularization term, i.e., Loss + Penalty, with respect to $X$ and $Y$. The loss component measures the distance between $R$ and its approximation $\hat{R}$, while the penalty term enforces model generalization and prevents overfitting. Accordingly, a latent factor model's optimization objective can be expressed as
\begin{equation}
\min_{X,Y}\; \mathcal{L}(X,Y)=\left\|A\odot\left(R-XY^{T}\right)\right\|_{F}^{2}+\lambda\left(\|X\|_{F}^{2}+\|Y\|_{F}^{2}\right),
\end{equation}
where $\odot$ indicates the Hadamard product that performs the element-wise multiplication between two matrices, $\lambda$ is a hyperparameter for adjusting the penalty effects, and $A$ is a $|U|\times|I|$ binary indexing matrix given as follows:
\begin{equation}
a_{u,i}=\begin{cases}
1, & r_{u,i}\in R_K,\\
0, & \text{otherwise},
\end{cases}
\end{equation}
where $r_{u,i}$ denotes the entry of $R$ at the $u$-th row and $i$-th column.

\subsection{Learning a Latent Factor Model via Gradient Descent Optimization}
Notably, since $R$ is extremely sparse, it is important to expand Eq. (1) into a density-oriented form as follows:
\begin{equation}
\mathcal{L}(X,Y)=\sum_{(u,i)\in R_K}\left(r_{u,i}-\sum_{d=1}^{f}x_{u,d}y_{i,d}\right)^2+
\lambda\left(\sum_{u=1}^{|U|}\sum_{d=1}^{f}x_{u,d}^{2}+\sum_{i=1}^{|I|}\sum_{d=1}^{f}y_{i,d}^{2}\right),
\end{equation}
where $x_{u,d}$ denotes the specific entry of $X$ at the $u$-th row and $d$-th column, and $y_{i,d}$ denotes the specific entry of $Y$ at the $i$-th row and $d$-th column, respectively.

The objective of Eq. (3) with $X$ and $Y$ can be optimized by gradient descent optimization. Then, a stochastic gradient descent algorithm can be adopted for optimizing Eq. (3). First, it needs to formulate the instant loss on a single entry $r_{u,i}$ in Eq. (3) as follows:
\begin{equation}
\varepsilon_{u,i}=\left(r_{u,i}-\sum_{d=1}^{f}x_{u,d}y_{i,d}\right)^2+
\lambda\left(\sum_{d=1}^{f}x_{u,d}^{2}+\sum_{d=1}^{f}y_{i,d}^{2}\right).
\end{equation}
Then, moving every single latent factor along the opposite direction of the stochastic gradient of Eq. (3) at the $n$-th learning iteration is expressed as follows:
\begin{equation}
\begin{aligned}
x_{u,d}^{n+1} &= x_{u,d}^{n}-\eta\frac{\partial \varepsilon_{u,i}^{n}}{\partial x_{u,d}^{n}},\\
y_{i,d}^{n+1} &= y_{i,d}^{n}-\eta\frac{\partial \varepsilon_{u,i}^{n}}{\partial y_{i,d}^{n}},
\end{aligned}
\end{equation}
where $\varepsilon_{u,i}^{n}$, $x_{u,d}^{n}$, and $y_{i,d}^{n}$ indicate the states of $\varepsilon_{u,i}$, $x_{u,d}$, and $y_{i,d}$ during the $n$-th iteration, and $\eta$ indicates the hyperparameter of learning rate, respectively. By combining Eqs. (4) and (5), the following learning strategies via gradient descent optimization can be obtained:
\begin{equation}
\begin{aligned}
e_{u,i}^{n} &= r_{u,i}-\sum_{d=1}^{f}x_{u,d}^{n}y_{i,d}^{n},\\
x_{u,d}^{n+1} &= x_{u,d}^{n}+\eta\left(e_{u,i}^{n}y_{i,d}^{n}-\lambda x_{u,d}^{n}\right),\\
y_{i,d}^{n+1} &= y_{i,d}^{n}+\eta\left(e_{u,i}^{n}x_{u,d}^{n}-\lambda y_{i,d}^{n}\right).
\end{aligned}
\end{equation}

\subsection{Learning a Latent Factor Model via Differential Evolution}
Differential evolution is a population-based method within the family of evolutionary algorithms \cite{ref19} that begins with a set of $NP$ individuals, each representing a potential solution to the target optimization problem. The algorithm evolves this population through three core operations: mutation, crossover, and selection. Based on this mechanism, we adopt differential evolution to optimize Eq. (1). Note that we introduce how to optimize $X$ by differential evolution, which is similar to optimizing $Y$.

First is the mutation operation. Although there are several mutation strategies, we chose DE/Rand/1 because of its efficiency and robustness \cite{ref19}. We initialize $NP$ latent factor matrices $X$. For each $X_{t,g}$, DE/Rand/1 generates its corresponding $M_{t,g}$ at the current generation $g$ as follows:
\begin{equation}
M_{t,g}=X_{t_1,g}+F\left(X_{t_2,g}-X_{t_3,g}\right),
\end{equation}
where $t_1,t_2,t_3\in\{1,2,\ldots,NP\}$ and $t_1\ne t_2\ne t_3\ne t$, $X_{t,g}$ denotes the $t$-th target latent factor matrix at the $g$-th generation, $M_{t,g}$ denotes the $t$-th mutation latent factor matrix at the $g$-th generation, and $F$ is the scaling factor for controlling the scaling of mutation operation. We have tested the other strategies and found that DE/Rand/1 performs better than the others overall.

Second is the crossover operation. Similarly, as discussed in \cite{ref19}, we choose the DE/CurrentToRand/1 strategy to generate the corresponding trial latent factor matrix $T_{t,g}$ for each $X_{t,g}$ at the current generation $g$ as follows:
\begin{equation}
T_{t,g}=X_{t,g}+K\left(M_{t,g}-X_{t,g}\right),
\end{equation}
where $K$ is a random number, $K\in[0,1]$, and $T_{t,g}$ denotes the $t$-th trial latent factor at the $g$-th generation. Next, bringing Eq. (7) into Eq. (8) and then simplifying it gives
\begin{equation}
T_{t,g}=(1-K)X_{t,g}+KX_{t_1,g}+KF\left(X_{t_2,g}-X_{t_3,g}\right).
\end{equation}
Third is the selection operation. After generating all the $T_{t,g}$ for each $X_{t,g}$, we need to decide which one between $T_{t,g}$ and $X_{t,g}$ should survive in the population at the next generation $g+1$:
\begin{equation}
X_{t,g+1}=\begin{cases}
T_{t,g}, & \mathcal{L}(T_{t,g},Y)\leq \mathcal{L}(X_{t,g},Y),\\
X_{t,g}, & \text{otherwise}.
\end{cases}
\end{equation}
After selection, the population evolved to generation $g+1$ generally maintains equal or improved performance relative to generation $g$ for the target optimization problem. Moreover, the scaling factor $F$ has a significant impact on both optimization effectiveness and convergence dynamics. The scale factor local search strategy is recognized as an effective approach for adaptively adjusting $F$ in differential evolution \cite{ref19}. Therefore, we adopt this strategy to dynamically determine the value of the scaling factor $F$ as follows:
\begin{equation}
F_i=\begin{cases}
F_l+\mathrm{random}_1\cdot F_u, & \mathrm{random}_2<\tau_1,\\
F_{\mathrm{SFGSS}}, & \mathrm{random}_2<\tau_2,\\
F_{\mathrm{SFHC}}, & \mathrm{random}_3<\tau_3,\\
F_i, & \text{otherwise},
\end{cases}
\end{equation}
where $\mathrm{random}_1$, $\mathrm{random}_2$, and $\mathrm{random}_3$ are uniform pseudo-random numbers between 0 and 1; $\sigma_1$, $\sigma_2$, and $\sigma_3$ are constant threshold values. Following \cite{ref19}, all hyperparameters in Eq. (11) are set as follows: $\mathrm{SFGSS}=8$, $\mathrm{SFHC}=20$, $F_l=0.1$, $F_u=0.9$, $\tau_1=0.1$, $\tau_2=0.03$, $\tau_3=0.07$, and $F_i$ is initialized to a random value between 0 and 1. Please refer to \cite{ref19} for more discussions regarding sensitivity of these hyperparameters.

\subsection{Customized Self-adaptive Weighting Mechanism}
To effectively ensemble the two diverse latent factor models optimized independently via differential evolution and gradient descent optimization, we employ the customized self-adaptive weighting mechanism to control their combination. Let $\mathcal{L}_{\mathrm{GDO}}$ be the partial loss optimized by gradient descent optimization, and $\mathcal{L}_{\mathrm{DE}}$ be the partial loss optimized by differential evolution at the $n$-th training iteration, respectively. $x^{\mathrm{GDO}}_{u,d}$ denotes the specific entry of $X^{\mathrm{GDO}}$ at the $u$-th row and $d$-th column, and $y^{\mathrm{GDO}}_{i,d}$ denotes the specific entry of $Y^{\mathrm{GDO}}$ at the $i$-th row and $d$-th column, respectively. Let $E_1^n$ and $E_2^n$ indicate the partial loss of gradient descent optimization and differential evolution at the $n$-th iteration, respectively. We make $\beta_1$ and $\beta_2$ control the combination of the two diverse latent factor models. The main principle is increasing $\beta_1$ and decreasing $\beta_2$ if $E_1^n<E_2^n$, and decreasing $\beta_1$ and increasing $\beta_2$ otherwise. At the $n$-th training iteration, $E_1^n$ and $E_2^n$ are computed as follows:
\begin{equation}
\begin{aligned}
E_1^n &= \sum_{(u,i)\in R_K}\left(r_{u,i}-\sum_{d=1}^{f}x_{u,d}^{\mathrm{GDO},n}y_{i,d}^{\mathrm{GDO},n}\right)^2,\\
E_2^n &= \sum_{(u,i)\in R_K}\left(r_{u,i}-\sum_{d=1}^{f}x_{u,d}^{\mathrm{DE},n}y_{i,d}^{\mathrm{DE},n}\right)^2.
\end{aligned}
\end{equation}
Let $C_1^n$ and $C_2^n$ be the cumulative loss corresponding to $E_1^n$ and $E_2^n$ until the $n$-th iteration, respectively. Then $C_1^n$ and $C_2^n$ are computed as follows:
\begin{equation}
C_j^n=\frac{\sum_{s=1}^{n}E_j^s}{\sum_{s=1}^{n}\left(E_1^s+E_2^s\right)},\quad j\in\{1,2\}.
\end{equation}
Assuming that $C_1^n$ and $C_2^n$ lie in the scale of $[0,1]$, then $\beta_1$ and $\beta_2$ are set in the following criterion:
\begin{equation}
\begin{aligned}
\beta_1^{n+1} &= \begin{cases}
\beta_1^n+\eta_{\beta}(1-\beta_1^n), & C_1^n<C_2^n,\\
\beta_1^n-\eta_{\beta}\beta_1^n, & C_1^n\ge C_2^n,
\end{cases}\\
\beta_2^{n+1} &= 1-\beta_1^{n+1},
\end{aligned}
\end{equation}
where $\beta_1^n$ and $\beta_2^n$ indicate the states of $\beta_1$ and $\beta_2$ at the $n$-th training iteration, and $\eta_\beta$ indicates a hyperparameter controlling their learning rates and can be set as suggested in \cite{ref18}. The convergence of such weighting mechanism of Eq. (14) is guaranteed. Please refer to \cite{ref18} for more details.

\subsection{Time Complexity Analysis}
ELFM-DEGDO's time complexity is decided by gradient descent optimization and differential evolution. The first part is $\Theta(N\times |R_K|\times f)$ and the second part is $\Theta(N\times G\times |R_K|\times f)$, where $N$ and $G$ denote the maximum number of iterations and generations.

\section{Experiments and Results}
\subsection{General Settings}
\textbf{Datasets.} Three HDI datasets are selected for evaluating the proposed ELFM-DEGDO model in the experiments and are summarized in Table~\ref{tab:datasets}.

\begin{table}[t]
\caption{Summary of experimental datasets.}
\label{tab:datasets}
\centering
\begin{tabular}{cccccc}
\toprule
No. & Name & $|U|$ & $|I|$ & $|R_K|$ & Density \\
\midrule
D1 & Eachmovie & 72,916 & 1,628 & 2,811,718 & 2.37\% \\
D2 & Flixter & 147,612 & 48,794 & 8,196,077 & 0.11\% \\
D3 & Jester & 24,983 & 100 & 1,186,324 & 47.49\% \\
\bottomrule
\end{tabular}
\end{table}

\textbf{Evaluation Metrics.} Missing data estimation is a common metric for representation learning for HDI data. Considering missing data estimation, the root mean squared error (RMSE) is widely used for evaluating estimation performance:
\begin{equation}
\mathrm{RMSE}=\sqrt{\frac{1}{|\Gamma|}\sum_{(u,i)\in\Gamma}\left(r_{u,i}-\hat{r}_{u,i}\right)^2},
\end{equation}
where $\Gamma$ denotes the testing set and $|\cdot|_{\mathrm{abs}}$ calculates the absolute value of a given number.

\textbf{Baselines.} We compare ELFM-DEGDO with several related models with different characteristics, including BLFM \cite{ref13}, FNLFM \cite{ref14}, L3FM \cite{ref18}, AutoRec \cite{ref15}, DCCR \cite{ref16}, and PMLFM \cite{ref17}.

\subsection{Comparison between ELFM-DEGDO and Baselines}
In this set of experiments, we assess the accuracy of missing data estimation. To ensure a fair comparison, the following configurations are used: (a) the latent factor dimension is fixed at $f=20$ for all LF-based methods, and (b) five-fold cross-validation is employed, with the average performance reported.

\begin{table*}[t]
\caption{The comparison results with win/loss counts and the Friedman test.}
\label{tab:comparison}
\centering
\small
\setlength{\tabcolsep}{4pt}
\resizebox{\textwidth}{!}{%
\begin{tabular}{ccccccccc}
\toprule
Dataset & Metric & BLFM & FNLFM & L3FM & AutoRec & DCCR & PMLFM & ELFM-DEGDO \\
\midrule
D1 & MAE & 0.1732 & 0.1763 & 0.1700 & 0.1784 & 0.1775 & 0.1728 & 0.1722 \\
 & RMSE & 0.2251 & 0.2259 & 0.2249 & 0.2305 & 0.2289 & 0.2245 & 0.2244 \\
D2 & MAE & 0.6447 & 0.6520 & 0.6318 & 0.6295 & 0.6308 & 0.6436 & 0.6407 \\
 & RMSE & 0.8961 & 0.9038 & 0.8960 & 0.8682 & 0.8792 & 0.8942 & 0.8945 \\
D3 & MAE & 0.7664 & 0.7778 & 0.7562 & 0.7905 & 0.7883 & 0.7652 & 0.7546 \\
 & RMSE & 0.9957 & 1.0003 & 0.9978 & 1.0078 & 1.0042 & 0.9944 & 0.9942 \\
\midrule
Statistical Analysis & win/loss & 6/0 & 6/0 & 4/2 & 4/2 & 4/2 & 5/1 & 29/7 \\
 & F-rank* & 4.5 & 5.67 & 3 & 5 & 4.7 & 3 & 2.17 \\
\bottomrule
\end{tabular}}
\end{table*}

The detailed results are shown in Table~\ref{tab:comparison}, from which it can be observed that ELFM-DEGDO achieves lower RMSE and MAE values compared to the competing approaches. To further verify whether ELFM-DEGDO significantly outperforms the other models in estimating missing data, statistical analyses are conducted. First, the penultimate row of Table~\ref{tab:comparison} shows the win/loss comparisons between ELFM-DEGDO and each baseline method, indicating that ELFM-DEGDO performs better in the majority of cases. Additionally, the Friedman test is applied, as it is a reliable non-parametric method for comparing multiple algorithms across different datasets. The corresponding results, reported in the final row of Table~\ref{tab:comparison}, reject the null hypothesis at the 0.05 significance level, confirming that there are statistically significant differences among the compared methods. Overall, the analysis demonstrates that ELFM-DEGDO provides superior accuracy in missing data estimation relative to the other models.

\subsection{Ablation Experiments}
To evaluate ELFM-GDO, ablation studies are performed by excluding each optimization component individually. The results are reported in Table~\ref{tab:ablation}, where ELFM-GDO represents the variant without differential evolution, while ELFM-DE excludes gradient descent optimization. The findings reveal that ELFM-DEGDO consistently delivers the strongest performance across all datasets with respect to both MAE and RMSE. For instance, on D1, ELFM-DEGDO achieves MAE and RMSE values of 0.1722 and 0.2244, respectively, surpassing both ELFM-GDO and ELFM-DE. Similar performance patterns can also be observed on D2 and D3. In general, the removal of either optimization strategy results in a noticeable decline in predictive accuracy, demonstrating that differential evolution and gradient descent jointly contribute to the effectiveness of ELFM-DEGDO.

\begin{table}[t]
\caption{The results of ablation experiments.}
\label{tab:ablation}
\centering
\setlength{\tabcolsep}{8pt}
\begin{tabular}{ccccc}
\toprule
Dataset & Metric & ELFM-GDO & ELFM-DE & ELFM-DEGDO \\
\midrule
D1 & MAE & 0.1733 & 0.1744 & 0.1722 \\
 & RMSE & 0.2252 & 0.2265 & 0.2244 \\
D2 & MAE & 0.6446 & 0.6451 & 0.6407 \\
 & RMSE & 0.8962 & 0.8989 & 0.8945 \\
D3 & MAE & 0.7665 & 0.7672 & 0.7546 \\
 & RMSE & 0.9956 & 0.9974 & 0.9942 \\
\bottomrule
\end{tabular}
\end{table}

\section{Conclusion}
This paper proposes ELFM-DEGDO for representing HDI data. The model integrates two complementary optimization strategies, differential evolution and gradient descent, and fuses them through a self-adaptive weighting mechanism to obtain more robust latent representations. Such ensemble is effective when handling heterogeneous HDI data. Results obtained from experiments on three real HDI datasets show that ELFM-DEGDO performs more effectively than multiple advanced latent factor models. We will improve ELFM-DEGDO by considering non-negativity constraints and boundary conditions.

\bibliographystyle{unsrt}
\bibliography{references}

\end{document}